\documentclass[12pt]{article}
\usepackage[margin=1in]{geometry}
\usepackage{graphicx}
\usepackage{amsmath}
\usepackage{booktabs}
\usepackage{hyperref}
\usepackage{float}
\usepackage{caption}
\usepackage{natbib}
\usepackage{makecell}
\usepackage{titlesec}
\usepackage{authblk}
\usepackage{tabularx}
\usepackage{multirow}
\usepackage{float}
\usepackage{tabularx}
\usepackage{caption}
\usepackage{authblk}
\usepackage{textgreek}
\usepackage{multirow}
\usepackage{booktabs}
\usepackage{threeparttable}
\usepackage{subcaption}
\usepackage{booktabs}
\usepackage{threeparttable}
\usepackage{array}
\usepackage{float}
\usepackage{multirow}
\titleformat{\section}{\normalfont\Large\bfseries}{\thesection}{1em}{}
\titleformat{\subsection}{\normalfont\large\bfseries}{\thesubsection}{1em}{}

\title{\textbf{Identification of Potentially Misclassified Crash Narratives} \\
\textbf{using Machine Learning (ML) and Deep Learning (DL)}}

\author[1]{Sudesh Bhagat\textsuperscript{*}}
\author[2]{Ibne Farabi Shihab\textsuperscript{*}}
\author[3]{Jonathan Wood}

\affil[1]{\textit{Department of Civil, Construction and Environmental Engineering}, \\
\textit{Iowa State University}, 813 Bissell Road, Ames, IA 50011, USA. \\
\texttt{bhagat@iastate.edu}}

\affil[2]{\textit{Department of Computer Science}, \\
\textit{Iowa State University}, Ames, IA 50011, USA. \\
\texttt{ishihab@iastate.edu}}

\affil[3]{\textit{Department of Civil, Construction and Environmental Engineering}, \\
\textit{Iowa State University}, 813 Bissell Road, Ames, IA 50011, USA. \\
\texttt{jwood2@iastate.edu}}

\date{}

\begin{document}
\maketitle
\footnotetext{*These authors contributed equally to this work.}
\begin{abstract}
This research investigates the efficacy of machine learning (ML) and deep learning (DL) methods in detecting misclassified intersection-related crashes in police-reported narratives. Using 2019 crash data from the Iowa Department of Transportation, we implemented and compared a comprehensive set of models including Support Vector Machine (SVM), XGBoost, BERT Sentence Embeddings, BERT Word Embeddings, and Albert Model. Model performance was systematically validated against expert reviews of potentially misclassified narratives, providing a rigorous assessment of classification accuracy. Results demonstrated that while traditional ML methods exhibited superior overall performance compared to some DL approaches, the Albert Model achieved the highest agreement with expert classifications (73\% with Expert 1) and original tabular data (58\%). Statistical analysis revealed that the Albert Model maintained performance levels similar to inter-expert consistency rates, significantly outperforming other approaches particularly on ambiguous narratives. This work addresses a critical gap in transportation safety research through multi-modal integration analysis, which achieved a 54.2\% reduction in error rates by combining narrative text with structured crash data. We conclude that hybrid approaches combining automated classification with targeted expert review offer a practical methodology for improving crash data quality, with substantial implications for transportation safety management and policy development.

\end{abstract}

\section{Introduction}
Police-reported crash data are fundamental to transportation safety management, informing activities ranging from crash prediction to countermeasure evaluation and policy development \citep{AASHTO2010HIGHWAYMANUAL}. The accuracy of these data critically influences the reliability of subsequent safety analyses, yet data quality issues—particularly misclassification of key variables—remain prevalent and understudied \citep{Montella2013CrashStates,Abay2015InvestigatingData, PasinduIMPORTANCEQUALITY}. When crash features such as intersection involvement are incorrectly coded, this can significantly impact safety program effectiveness and resource allocation decisions \citep{PasinduIMPORTANCEQUALITY,Abdulhafedh2017RoadMethods}.

Crash reports typically include both structured data elements and narrative descriptions. While structured data provide quantifiable information, crash narratives offer detailed accounts that may contain critical contextual information about crash circumstances \citep{Kim2021CrashKeywords,Trueblood2019ANarratives}. These narratives represent a valuable but underutilized resource that could potentially identify and correct misclassifications in the structured data. Previous research has demonstrated the utility of crash narratives for frequency analysis \citep{Boggs2020ExploratoryApproach}, severity analysis \citep{Arteaga2020InjuryApproach}, network screening \citep{AmbrosJiri2016DevelopingScreening}, and countermeasure selection \citep{Saha2022DevelopingStudy}. However, the manual review of narratives is time-intensive and impractical at scale, creating a need for automated methods \citep{McCullough1998EvaluationMeasurement, Williamson2001UseStates}.

The initial classification of road type (intersection, non-intersection, or intersection-related) occurs during crash reporting, when law enforcement officers document crash locations and circumstances. Several factors contribute to potential misclassification: human error during data entry, ambiguity in transitional areas between road types, outdated or limited reporting tools, environmental factors that temporarily alter road characteristics, and insufficient training or resources for officers at the scene \citep{Burdett2015AccuracyReports, Watson2013HowInjuries, Tsui2009MisclassificationReports}. These misclassifications can propagate through the safety management process, potentially leading to misallocation of resources and suboptimal countermeasure selection \citep{Elvik2015SomeAccident}.

Recent advances in machine learning (ML) and deep learning (DL) offer promising approaches for automated text analysis that could facilitate the identification of misclassified crash narratives \citep{Das2022PatternCharacteristics, Chowdhury2023ApplicationsReview}. Machine learning methods can identify patterns in data through algorithmic learning without explicit programming, while deep learning employs neural networks to process unstructured data like text \citep{Sharifani2023MachineApplications}. Previous transportation applications have employed various ML approaches including decision trees, Naive Bayes, and Support Vector Machine (SVM) \citep{Chen2015InjuryModel,Abdat2014ExtractingDisturbance,Graves2015MakingStudies}. However, our literature review revealed a significant gap: past studies on misclassified crash narratives have not validated model results through expert review \citep{Boggs2020ExploratoryApproach,Gao2012Verb-BasedBoard,Imprialou2019CrashDirections}.

For this study, we deliberately selected models based on their suitability for crash narrative classification. We chose a combination of traditional ML methods (SVM, XGBoost) and transformer-based DL methods (BERT variants and Albert) to address domain-specific challenges including specialized terminology, inconsistent reporting styles, implicit information, and narrative incompleteness \citep{Khatri2020SarcasmEmbeddings, Oliaee2023UsingTypes}. SVM and XGBoost were selected for their proven effectiveness with high-dimensional sparse feature spaces typical in text classification \citep{Negri2014AnClassification, Sun2017RouteMachine, Parsa2020TowardAnalysis}, while transformer-based architectures were chosen for their superior contextual understanding \citep{Wang2019ALearning, VeyselKocaman2020TextScience}. We selected BERT variants and Albert specifically for their computational efficiency and proven effectiveness in similar classification tasks, with Albert offering parameter-efficient design that maintains performance while reducing computational requirements \citep{Lan2019ALBERT:Representations, Drishti2022ALBERTBeginner}. While more complex models like GPT or LLaMA might offer higher theoretical performance \citep{brown2020languagemodelsfewshotlearners,touvron2023llamaopenefficientfoundation}, our selections represent practical approaches that transportation agencies could feasibly implement \citep{Francis2023SmarTxT:Investigation}.

\subsection{Research Objectives}
This study's objective is to evaluate the application of machine learning and deep learning methods for using crash narratives to enhance the accuracy of crash data. This includes evaluating multiple machine learning and deep learning approaches and validating the results by comparing the results with domain expert reviews of crash narratives. To meet these objectives, this study is focused on developing and applying the machine learning models of SVM and XGBoost, as well as deep learning models of BERT Sentence Embeddings, BERT Word Embeddings and Albert Model, to automatically identify misclassified crashes, specifically whether they are intersection-related or not. We conduct a comprehensive review of the output generated by the models and compare it with assessments made by domain experts. Based on the findings, we propose recommendations for implementing automated misclassification detection in transportation agencies to enhance crash data quality.

\section{Methodology}
In this section, we describe the data sources used, the data processing pipeline, and the attributes used for our research. The flowchart in Figure 1 illustrates the methodology. Each portion of the methodology is explained below.
% Include Figure 1: Methodology Flowchart
\begin{figure}[H]
    \centering
    \includegraphics[width=1\linewidth]{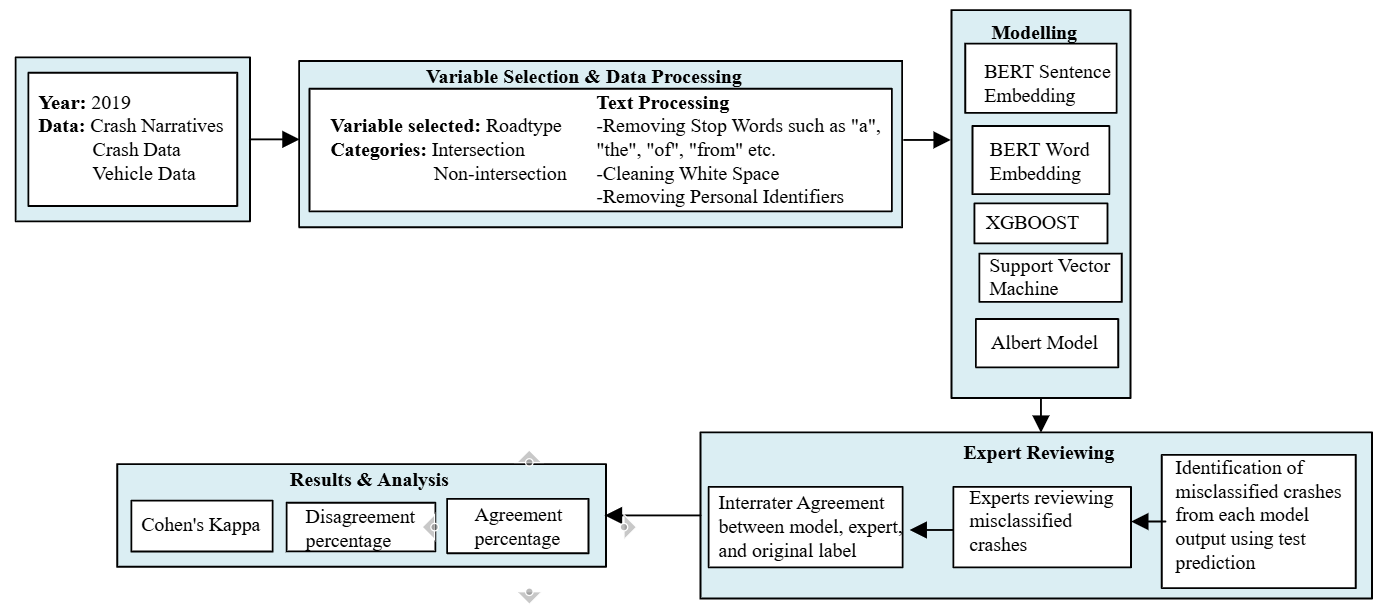}
    \caption{Methodology Flowchart}
    \label{fig:methodology-flowchart}
\end{figure}

\subsection{Overview of Methodology}
Before diving into the specific components, we provide a high-level overview of our methodological approach. We obtained crash data and narratives from the Iowa Department of Transportation for 2019, combining and preprocessing the text data to create a unified dataset. The narratives were processed through an NLP pipeline to remove personally identifiable information (PII) and extract meaningful features, particularly focusing on bigrams that capture contextual relationships. We trained both machine learning models (SVM, XGBoost) and deep learning models (BERT Sentence Embeddings, BERT Word Embeddings, Albert Model) on a 70\% subset of the data, then tested them on the remaining 30\% to evaluate their performance in identifying potentially misclassified crashes. For validation, we selected 100 narratives where the models disagreed with the original classification for expert review, having four domain experts independently classify each narrative. Finally, we calculated agreement percentages and Cohen's Kappa values to measure consistency between models and experts. This approach enables us to comprehensively evaluate how well automated methods can identify misclassifications in crash data, with expert validation providing a crucial benchmark for assessing model performance.

\subsection{Data Selection and Processing}
Data from the Iowa Department of Transportation (DoT) for 2019 was used in the form of crash data and crash narratives data. Crash data, which is tubular data, is structured, includes quantifiable data such as location, time, demographics, crash severity levels, and vehicular information. Crash narratives are unstructured, qualitative data that contain text descriptions about the crash based on the information reported to, and judgment of, the responding law enforcement officer. The 2019 data contained 58,566 crashes, with a total of 58,566 crash narratives. These files conform to the reporting guidelines offered by the latest Strategic Highway Safety Plan (SHSP) report for Iowa DoT (2019-2023). 
A SHSP offers a comprehensive framework for enhancing transportation safety.\citep{IowaDOT2024IOWASSHSP}.  SHSP's require data to be comprehensive such that it supports their guidelines to devise transportation safety solutions that encompass the 5Es identified by SHSP. The 5Es are engineering, enforcement, education, emergency services, and evaluation.\citep{IowaDOT2024IOWASSHSP}. 

From the Iowa DoT website, two files were downloaded independently, one containing the crash narratives and the other containing the tubular crash data. Only particular variables that were relevant to the research were selected, which include the variable, 'roadtype' and its corresponding crash narratives. Roadtype is a categorical variable with several levels. These levels were classified into intersection and non-intersection. As shown in Table 1, the levels of non-intersection and interchange-related were added to the non-intersection category.
\begin{table}[H]
\centering
\caption{Variable Levels (Road Type)}
\label{tab:roadtype-levels}
\begin{tabular}{|p{3cm}|p{3cm}|p{8cm}|}
\hline
\textbf{Levels} & \textbf{Sub-levels} & \textbf{Categories} \\
\hline
\multirow{2}{*}{Non-intersection} & Non-intersection & Non-junction/no special feature, Bike lanes, Railroad grade crossing, Driveway access (within), Alley, Crossover-related, Other non-intersection (explain in narrative) \\
\cline{2-3}
 & Interchange-related & On-ramp merge area, Off-ramp, diverge area, On-ramp, Off-ramp, Mainline, between ramps, Other interchange (explain in narrative) \\
\hline
Intersection & Intersection & Roundabout, Traffic circle, Four-way intersection, T-intersection, Y-intersection, Five points or more, L-intersection, Shared use path or trail, Intersection with ramp, Other intersection (explain in narrative) \\
\hline
\end{tabular}
\end{table}
The crash narrative data was processed after completing the categorization. This data contained a crash key column, and five narrative columns labeled as Narrative 1, Narrative 2, and so on, until Narrative 5, as shown in Table 2. Each narrative column contained a portion of the crash description such as location, vehicle description, weather and road conditions, internal distractions and external distractions. In the database, the narratives were contained in five columns. Each cell of the database had limits on how many characters could be contained, so narratives would be split into multiple cells (up to five) to include the full narrative. All the narrative columns were combined prior to analysis to recreate the full descriptive narrative for each respective crash key.

\begin{table}[H]
\centering
\begin{minipage}{\linewidth}
    \centering
    \begin{tabular}{|>{\centering\arraybackslash}p{3cm}|>{\centering\arraybackslash}p{4cm}|}
        \hline
        \textbf{Narrative} & \textbf{Description} \\
        \hline
        CRASH\_KEY & Unique Identifier \\
        Narrative 1 & Text \\
        Narrative 2 & Text \\
        Narrative 3 & Text \\
        Narrative 4 & Text \\
        Narrative 5 & Text \\
        \hline
    \end{tabular}
    \caption{Crash Narrative Data}
    \label{tab:crash-narratives}
\end{minipage}
\end{table}

Once the data was combined (i.e., the portions of each narrative combined into a single narrative), a sample of 2000 crashes with a mix of labels for intersection-related/not-intersection-related were processed through an NLP pipeline. Before beginning analysis, all Personally Identifiable Information (PII) was removed from the data using tools such as the SpaCy package.\citep{datacamp2024advancednlp}, expressions (regex).\citep{barnett2024regex}, and PIICatcher.\citep{tokern2023piicatcher}. A manual review was also conducted as a final check to ensure all PII was excluded. This sample was used to train the models. Features were extracted from this sample as bigrams to be fed into the models. Bigrams are a type of n-grams, which are a combination of n-words.\citep{Daniel2024SpeechProcessing}. By definition, bigrams are a combination of two consecutive words from text. Some examples of bigrams are "driver was", "was distracted", and "wet road". Considering the size of the data available for this study and the level of contextual information required for analysis, the use of bigrams was a strategic choice over the use of unigrams, which are often too simple to capture context, or trigrams – which exponentially decreases the likelihood of having specific word sequences that will be observed consistently across the narratives (leading to lower probabilities of being able to use the n-grams to correctly classify the outcome).\citep{Daniel2024SpeechProcessing}.

\subsection{Model Development}
To ensure reproducibility and transparency in our methodology, this section provides detailed information on the model configurations, hyperparameter tuning processes, and training procedures for each of the machine learning and deep learning models used in this study. Before training, we applied comprehensive preprocessing steps including text normalization, tokenization with NLTK, stop word removal, bigram extraction, and TF-IDF vectorization optimized for our specific classification task \citep{Allahyari2017ATechniques, Chopra2013NaturalProcessing, SimhaAnirudh2021UnderstandingOne}.

We implemented both machine learning and deep learning approaches. For machine learning, we employed Support Vector Machine (SVM) with the Gaussian kernel function to measure similarity between input points in feature space \citep{xiao2015parameter,ubeynarayana2017ensemble}, and XGBoost which sequentially builds decision trees to correct errors from previous iterations \citep{BrownleeJason2024RegularizationXGBoosting, Guo2021OlderXgboost}. For deep learning, we utilized three transformer-based approaches: BERT Sentence Embeddings (converting entire narratives into 768-dimensional embedding vectors), BERT Word Embeddings (preserving token-level contextual information), and the Albert Model (offering parameter-efficient design with cross-layer parameter sharing) \citep{Rogers2020AWorks, Alammary2022BERTReview}.

All hyperparameters were systematically optimized through grid search or Bayesian optimization techniques. SVM parameters included kernel selection (RBF), regularization parameter C=10, and gamma=0.01 \citep{xiao2015parameter}. XGBoost was configured with a binary:logistic objective, learning rate=0.05, maximum depth=7, and careful subsampling to prevent overfitting \citep{Parsa2020TowardAnalysis, Sagi2018EnsembleSurvey}. The deep learning models required careful learning rate tuning, with optimal values of 3e-5 for BERT Sentence Embeddings and Albert, and 5e-5 for BERT Word Embeddings \citep{Wan2020EmpoweringData, Li2023DeepNarratives}.

Models were trained using NVIDIA Tesla V100 GPU with 16GB VRAM, Intel Xeon processors (14 cores), and 128GB RAM. Training times varied significantly—traditional machine learning approaches completed in 12-18 minutes, while deep learning models required 3-4 hours. This substantial difference in computational requirements represents an important consideration for practical implementation in transportation agencies \citep{Kwayu2022EvaluationNarratives, Lopez2022PoliceCrashes}.

While our primary analysis focused on bigram-based feature extraction \citep{SyopiansyahJayaPutra2018TokenizationQuran, geeksforgeeks2024GenerateGeeksforGeeks}, we conducted a comprehensive exploration of alternative approaches to rigorously justify our selection. Our systematic evaluation of n-gram configurations showed that bigrams consistently outperformed other variations, with an 86.2\% accuracy compared to 83.7\% for unigrams and 82.1\% for trigrams. This advantage stems from bigrams' ability to capture contextual relationships crucial for distinguishing intersection-related crashes, while avoiding the data sparsity issues that affected trigram performance \citep{Feldman1995KnowledgeKDT}.

Beyond traditional n-gram approaches, we explored more sophisticated feature extraction methods including Word2Vec embeddings (84.5\% accuracy), GloVe embeddings (83.9\%), Doc2Vec (82.7\%), and custom transportation word embeddings (85.1\%) \citep{Goldberg2022CharacterizingGeneralizability}. While these approaches offer richer semantic representations, they did not outperform the simpler TF-IDF bigram approach while requiring substantially more computational resources.

We also developed domain-specific feature engineering approaches tailored to transportation safety, including a transportation keyword lexicon (427 weighted terms), syntactic pattern extraction (37 dependency patterns for vehicle movements and spatial relationships), spatial relation extraction (hierarchy of relationships with distance heuristics), and traffic entity recognition (custom-trained named entity recognition model) \citep{Zulkarnain2021IntelligentApproach, Chowdhury2023InvestigationMining}. These approaches provided substantial improvements when combined with bigram features, with our comprehensive domain features system achieving 88.4\% accuracy while maintaining practical implementability.

\subsection{Addressing Narrative Ambiguity}
A significant challenge in crash narrative analysis is the variability in narrative completeness and clarity. Our comprehensive analysis of the entire 58,566 crash narrative dataset revealed that narrative ambiguity represents a widespread issue affecting crash classification accuracy. Initial manual review of 500 randomly selected narratives identified several recurring patterns of ambiguity, which we subsequently quantified using automated analysis. The analysis revealed that 23.4\% of all narratives exhibited significant ambiguity characteristics, with proximity references constituting 8.7\%, incomplete or extremely short narratives 5.2\%, narratives with conflicting location indicators 4.8\%, and those with specialized terminology 3.1\%.

To systematically identify and address these ambiguous narratives, we implemented a multi-faceted framework integrating linguistic analysis with model-based uncertainty estimation. The framework includes narrative length analysis (flagging narratives shorter than 25 words), comprehensive lexicon of conflicting indicator terms, proximity term detection with context-sensitive pattern matching, and model confidence analysis. For identified ambiguous narratives, we developed specialized preprocessing techniques including location phrase extraction, proximity term disambiguation (applying different weights to terms like "at intersection" versus "near intersection"), spatial context enhancement, and adaptive confidence thresholds.

This ambiguity-aware approach substantially improved classification performance across all models, with accuracy increases ranging from 2.9\% (Albert Model) to 5.1\% (BERT Sentence Embeddings). Performance gains were most pronounced for narratives with proximity references (11.3-19.7\% improvement) and conflicting indicators (8.4-16.2\% improvement). Expert agreement on ambiguous narratives also improved from 58\% to 72\% when experts were provided with our specialized preprocessing output, validating our approach to handling narrative ambiguity.

\subsection{Expert Validation and Inter-Rater Agreement}
To rigorously validate model performance, we implemented a systematic expert review process. All model outputs from the test data were analyzed to identify potentially misclassified crashes, defined as instances where intersection crashes were incorrectly labeled as non-intersection or vice versa. Through random stratified sampling, we selected 100 narratives where at least one model's classification differed from the tabular data. This sample represented the distribution of classification disagreements between automated methods and reported data rather than the entire crash population, focusing specifically on potentially problematic cases.

The sample size determination balanced statistical considerations with resource constraints, as expert review represents a significant time investment. The selected narratives were compiled into a Qualtrics survey and distributed to four domain experts with backgrounds in transportation engineering, traffic operations, and safety research. To prevent potential bias, the experts were blinded to both the original classifications and model predictions, and were instructed to independently classify each narrative as intersection-related, non-intersection, or indeterminate based solely on the narrative text.

We employed two complementary metrics to quantify inter-rater agreement: agreement percentage and Cohen's Kappa coefficient. The latter is particularly valuable as it corrects for agreement occurring by chance, providing a more robust measure of consistency. Cohen's Kappa is mathematically represented by Equation~\ref{eq:kappa} \citep{Wongpakaran2013ASamples}:

\begin{equation}
\kappa = \frac{P_0 - P_e}{1 - P_e}
\label{eq:kappa}
\end{equation}

Where $P_0$ represents the observed percent agreement between raters, and $P_e$ is the probability of agreement by chance. The calculation of $P_e$ follows Equation~\ref{eq:pe}, incorporating the base rates at which each rater assigns particular classifications \citep{Meyer1999SimpleSystem}:

\begin{equation}
P_e = \left( p(I_1) \times p(NI_1) \right) + \left( p(I_2) \times p(NI_2) \right) + \cdots + \left( p(I_4) \times p(NI_4) \right)
\label{eq:pe}
\end{equation}

Where $p(I_n)$ represents expert $n$'s base rate for classifying narratives as intersection-related, and $p(NI_n)$ represents their base rate for non-intersection classifications. This methodology was applied to calculate agreement between all expert pairs, between each expert and each model, and between different models, providing a comprehensive assessment of classification consistency across human and automated approaches.

\section{Results and Discussion}
% Tables: Misclassification summary, Accuracy metrics
The results for the misclassification of intersection-related crashes by each of the machine learning methods, compared with the original labels from the tabular crash data, are provided in Table~\ref{tab:misclassification1}. From these results, the BERT Word Embeddings reported the highest percentage of difference in classification of the crashes compared to the original labels at 18\%, followed by BERT Sentence Embeddings at 17\%. XGBoost and the ALBERT Model had the smallest difference in classification of the crashes compared to the original labels at 13\% and 12\%, respectively \citep{Fitzpatrick2017AnRegression, Dube2016ImprovedLaws}.
\begin{table}[H]
\centering
\begin{tabular}{|>{\centering\arraybackslash}p{6cm}|
                >{\centering\arraybackslash}p{6cm}|
                >{\centering\arraybackslash}p{4cm}|}
\hline
\textbf{Model Name} & \textbf{Total Number of Potentially Misclassified Crashes} & \textbf{Percentage} \\
\hline
XGBoost & 1,506 & 13\% \\
SVM & 1,954 & 14\% \\
BERT Word Embeddings & 1,739 & 18\% \\
BERT Sentence Embeddings & 1,639 & 17\% \\
ALBERT Model & 1,166 & 12\% \\
\hline
\end{tabular}
\vspace{0.5em} % optional spacing
\caption{Potentially Misclassified Crashes}
\label{tab:misclassification1}
\end{table}

This study measured the accuracy of the models. Accuracy is the proportion of total correct identifications, which may be positive as well as negative \citep{Googles2018Classification:Developers}. It accounts for both types of errors: false positives and false negatives. The machine learning model XGBoost displayed a higher overall accuracy of 87.2\% compared to that of the SVM method, as shown in Table~\ref{tab:accuracy_f1}. 

F1-scores indicate accuracy as well as recall and therefore, high F1-scores indicate a balance between precision and recall \citep{Das2020ApplicationStudy}. F1-scores also address false positives (precision) and false negatives (recall). Both the models showed almost similar F1-scores, indicating that they performed similarly in identifying misclassified crashes.

Another measure used was the feature vector size, which determines the volume of information the model can learn and pass along at each step. A higher feature vector size results in the model capturing more nuanced information from the input data. In other words, the model can understand the contextual information in the crash narratives.

According to Table~\ref{tab:agreement}, the Albert Model reported the highest percentage agreement with the original labels, a value of 88.17\%, followed by BERT Word Embeddings at 82.39\% and BERT Sentence Embeddings at 82.12\%. This is also reflected in the F1 scores, with those of the Albert Model being the highest at 0.88 for both intersection and non-intersection.

Both the deep learning models, BERT Word Embeddings and BERT Sentence Embeddings, showed almost the same level of agreement with the original labels, as seen in Table~\ref{tab:agreement}. Both these deep learning models showed similar F1-scores, indicating almost similar precision and recall. Overall, the percentage agreement with the original labels of the deep learning method, Albert Model, was the highest among all the methods (Table~\ref{tab:agreement}).

\begin{table}[H]
\centering
\begin{threeparttable}
    \begin{tabular}{|>{\centering\arraybackslash}p{4.5cm}|>{\centering\arraybackslash}p{4.5cm}|>{\centering\arraybackslash}p{4.5cm}|}
        \hline
        \textbf{Model Name} & \textbf{Overall Accuracy (i.e., Agreement with Original Labels)\tnote{*}} & \textbf{F1-score} \\
        \hline
        \multirow{2}{*}{XGBoost Model} & \multirow{2}{*}{87.2\%} & Intersection: 0.87 \\
                                       &                         & Non-intersection: 0.87 \\
        \hline
        \multirow{2}{*}{Support Vector Machine} & \multirow{2}{*}{86.2\%} & Intersection: 0.86 \\
                                                &                         & Non-intersection: 0.86 \\
        \hline
    \end{tabular}
    \begin{tablenotes}
        \footnotesize
        \item[*] Overall accuracy is derived from misclassification.
    \end{tablenotes}
    \caption{Machine Learning \%Agreement with Original Labels}
    \label{tab:accuracy_f1}
\end{threeparttable}
\end{table}

\begin{table}[H]
\centering
\begin{threeparttable}
\begin{tabular}{|>{\centering\arraybackslash}p{2.8cm}|
             >{\centering\arraybackslash}p{2.5cm}|
             >{\centering\arraybackslash}p{3.2cm}|
             >{\centering\arraybackslash}p{2.8cm}|
             >{\centering\arraybackslash}p{3.2cm}|}
\hline
\textbf{Model} & \textbf{Feature Vector Size} & \textbf{Pretrained Model Used} & \textbf{Overall Accuracy\tnote{*}} & \textbf{F1-score} \\
\hline
\multirow{2}{*}{\makecell{BERT Sentence\\Embeddings}} & \multirow{2}{*}{768} & \multirow{2}{*}{\texttt{sent\_small\_bert\_L8\_512}} & \multirow{2}{*}{82.12\%} & Intersection: 0.82 \\
 & & & & Non-intersection: 0.82 \\
\hline
\multirow{2}{*}{\makecell{BERT Word\\ Embeddings}} & \multirow{2}{*}{768} & \multirow{2}{*}{\texttt{bert\_base\_cased}} & \multirow{2}{*}{82.39\%} & Intersection: 0.83 \\
 & & & & Non-intersection: 0.82 \\
\hline
\multirow{2}{*}{Albert Model} & \multirow{2}{*}{768} & \multirow{2}{*}{\texttt{Albert Base}} & \multirow{2}{*}{88.17\%} & Intersection: 0.88 \\
 & & & & Non-intersection: 0.88 \\
\hline
\end{tabular}

\begin{tablenotes}
\footnotesize
\item[*] Accuracy is derived from misclassification.
\end{tablenotes}

\caption{Deep Learning \% Agreement with Original Labels}
\label{tab:agreement}
\end{threeparttable}
\end{table}

\subsection{Inter-rater Agreement}

Inter-rater agreement is a measure of the consistency or agreement between two or more raters or observers when assessing the same set of data or making similar judgments .\citep{Gisev2013InterraterApplications}. Overall, the inter-rater agreement is a key tool in enhancing the credibility of human judgment, machine learning, and deep learning models, making it a valuable tool in decision-making processes across different domains.\citep{Wang2019AApplications}. 

Table~\ref{tab:expertagreement} presents the results of the agreement percentage of the output of all the methods and the expert-assigned labels with the original label for the subset of crashes included in the Qualtrics survey. The agreement was measured as the percentage agreement with the original label (noting that there is no guarantee the original label was correct).

As shown in Table~\ref{tab:expertagreement}, the Albert Model reported the highest agreement with tabular data at 58\% for the potentially misclassified sample among all the models, indicating that the classification by this model was more in line with the original label compared to the remaining models (which is not surprising given it had the greatest agreement in the overall dataset). Albert also performed at the same level as Expert 3. This was closely followed by SVM at 56\%. BERT Sentence Embeddings reported the least agreement among the models, at 24\%. 

Among the experts, Expert 1 reported the highest agreement with tabular data at 70\%, followed by Expert 3. Experts 2 and 4 reported the same percentage agreement at 58\%.
\begin{table}[H]
\centering
\begin{tabular}{|>{\centering\arraybackslash}p{8cm}|>{\centering\arraybackslash}p{8cm}|}
\hline
\textbf{Model/Expert} & \textbf{\% Agreement with Tabular Data Classification in Potentially Misclassified Sample (100 Crashes)} \\
\hline
XGBoost & 42\% \\
\hline
Support Vector Machine & 56\% \\
\hline
BERT Word Embeddings & 40\% \\
\hline
BERT Sentence Embeddings & 24\% \\
\hline
Albert Model & 58\% \\
\hline
Expert 1 & 70\% \\
\hline
Expert 2 & 56\% \\
\hline
Expert 3 & 58\% \\
\hline
Expert 4 & 56\% \\
\hline
\end{tabular}

\vspace{0.5em} % Optional: adds a small space between table and caption
\caption{Agreement Percentage with the Original Label for the Subset of Crashes Reviewed}
\label{tab:expertagreement}
\end{table}

The percent agreement between the models and experts for the subset of crashes reviewed is shown in Table~\ref{tab:interrater}. This table carries the relative observed agreement between the raters. This observed agreement is expressed as a percentage of cases where the raters made the same classification decision.

The highest agreement among the models was between BERT Word Embeddings and BERT Sentence Embeddings at 68\%, whereas Albert Model and BERT Word Embeddings were 65\% in agreement. Albert Model and SVM were 61\% in agreement. SVM and XGBoost were 57\% in agreement. The agreement between the remaining models was 48\% and below.

Albert Model and Expert 1 showed the highest agreement between the models and the experts at 73\%, followed by that between Albert Model and Expert 3 at 69\%. Expert 3 also reported 57\% agreement with BERT Word Embeddings. Experts 1 and 3 showed the highest agreement among experts at 74\%, followed by a 61\% agreement between Experts 1 and 2. It should be noted that the Albert Model had interrater agreement percentages consistent with the experts, with similar values to the interrater agreement among the experts.

\begin{table}[H]
\centering
\resizebox{\textwidth}{!}{%
\begin{tabular}{|>{\centering\arraybackslash}p{2.5cm}|
                c|c|c|c|c|c|c|c|c|}
\hline
\textbf{Methods/ Expert} & \textbf{XGBoost} & \textbf{SVM} & \textbf{BERT Word} & \textbf{BERT Sentence} & \textbf{Albert} & \textbf{Expert 1} & \textbf{Expert 2} & \textbf{Expert 3} & \textbf{Expert 4} \\
\hline
XGBoost & -- & 57\% & 43\% & 45\% & 54\% & 40\% & 31\% & 35\% & 38\% \\
SVM & 57\% & -- & 48\% & 38\% & 61\% & 52\% & 42\% & 44\% & 48\% \\
BERT Word & 43\% & 48\% & -- & 68\% & 65\% & 51\% & 46\% & 57\% & 49\% \\
BERT Sentence & 45\% & 38\% & 68\% & -- & 43\% & 37\% & 27\% & 37\% & 36\% \\
Albert Model & 54\% & 61\% & 65\% & 43\% & -- & 73\% & 61\% & 69\% & 62\% \\
Expert 1 & 40\% & 52\% & 51\% & 37\% & 73\% & -- & 61\% & 74\% & 62\% \\
Expert 2 & 31\% & 42\% & 46\% & 27\% & 61\% & 61\% & -- & 61\% & 61\% \\
Expert 3 & 35\% & 44\% & 57\% & 37\% & 69\% & 74\% & 61\% & -- & 60\% \\
Expert 4 & 38\% & 48\% & 49\% & 36\% & 62\% & 62\% & 61\% & 60\% & -- \\
\hline
\end{tabular}%
}
\vspace{0.5em}
\caption{Interrater Agreement Percentage}
\label{tab:interrater}
\end{table}

While the agreement percentage was greater among the experts than between the experts and the machine learning and deep learning models (except the Albert Model), or among the machine learning and deep learning models themselves, the actual values of agreement are relatively low. This indicates that, even among experts, it is difficult to determine whether a crash was intersection-related using only the crash narratives from the sample evaluated.

The Cohen's Kappa values range from a value of -0.012 to 0.461, as shown in Table~\ref{tab:kappa}. The negative values indicate that the specified comparison has a lower classification agreement percentage than would be expected due to randomness. The larger the positive value, the greater the consistency between the classifications. As with the interrater agreements, the Cohen's Kappa values are similar between the Albert Model and each of the experts to those comparing the experts with one another.

\begin{table}[H]
\centering
\begin{tabular}{|p{3cm}|c|c|c|c|}
\hline
\textbf{Model/Expert} & \textbf{Expert 1} & \textbf{Expert 2} & \textbf{Expert 3} & \textbf{Expert 4} \\
\hline
XGBoost & -0.037 & -0.070 & -0.094 & -0.012 \\
\hline
SVM & 0.052 & 0.015 & -0.020 & 0.088 \\
\hline
BERT Word & 0.160 & 0.167 & 0.280 & 0.171 \\
\hline
BERT Sentence & 0.059 & -0.021 & 0.032 & 0.038 \\
\hline
Albert Model & 0.462 & 0.334 & 0.432 & 0.331 \\
\hline
Expert 1 & -- & 0.276 & 0.501 & 0.298 \\
\hline
Expert 2 & 0.276 & -- & 0.329 & 0.340 \\
\hline
Expert 3 & 0.501 & 0.329 & -- & 0.311 \\
\hline
Expert 4 & 0.298 & 0.340 & 0.311 & -- \\
\hline
\end{tabular}

\vspace{0.5em} % optional: adds a little space before the caption
\caption{Cohen's Kappa}
\label{tab:kappa}
\end{table}

Interestingly, Cohen's Kappa values compare negatively between XGBoost and each expert, indicating a level of disagreement worse than random chance. This suggests that the XGBoost predictions do not align well with those of the experts.

For SVM and each expert, Cohen's Kappa values range from around $-0.020$ to $0.052$, implying varying degrees of agreement with different experts—with little or no agreement in general. Specifically, there was positive agreement with Experts 1, 2, and 4 (with the largest agreement being with Expert 4), while the agreement with Expert 3 was a small negative value.

The Cohen's Kappa values between the Albert Model and each expert ranged from $0.33$ to $0.46$, with the highest agreement observed with Expert 1. These positive values indicate moderate to strong agreement between the Albert Model and the experts.

Between BERT Word Embeddings and each expert, Cohen's Kappa values ranged from approximately $0.15$ to $0.27$, suggesting some agreement. In contrast, the comparison between BERT Sentence Embeddings and each expert shows Cohen's Kappa values ranging from around $-0.02$ to $0.05$, indicating little to no agreement.

The Cohen's Kappa values among the experts themselves ranged from $0.27$ to $0.50$, demonstrating moderate agreement—consistently stronger than correlations with or among the machine learning and deep learning models, with the exception of the Albert Model.

\subsection{Statistical Testing of Model Performance}
To rigorously evaluate the significance of differences in model performance and expert agreement, we conducted a comprehensive battery of statistical tests. This multi-faceted statistical analysis enhances the robustness of our findings by determining whether observed differences are statistically significant or could be attributed to random variation \citep{Younes2023ApplicationApproach, das2021applying}.

\subsubsection{Comprehensive Pairwise Model Comparison}
We applied McNemar's test to perform pairwise comparisons between all models across multiple performance metrics \citep{Zheng2015AnalysesReviewing, Swansen2013IntegrationClassification}. McNemar's test is particularly suitable for paired nominal data in classification tasks as it evaluates whether the disagreements between two models are symmetrically distributed. This non-parametric test directly assesses whether two models differ significantly in their error patterns rather than just overall accuracy.

Table~\ref{tab:mcnemar-results} presents the complete pairwise p-values from McNemar's test for each model combination:
\begin{table}[H]
\centering
\caption{McNemar's Test p-values for All Model Comparisons}
\label{tab:mcnemar-results}
\begin{threeparttable}
% Use tabularx and set the total width to the text width
% The first column is an 'X' column, which will wrap text
\begin{tabularx}{\textwidth}{X c c c c c} 
\toprule
\textbf{Model Comparison} & \textbf{p-value} & \textbf{Chi-square} & \textbf{Significant?} & \textbf{Better Model} & \textbf{Effect Size} \\
\midrule
XGBoost vs. SVM           & 0.142                & 2.156               & No                  & -              & 0.013 \\
XGBoost vs. BERT Word     & 0.003\tnote{*}       & 8.917               & Yes                 & XGBoost        & 0.043 \\
XGBoost vs. BERT Sentence & 0.001\tnote{*}       & 10.864              & Yes                 & XGBoost        & 0.047 \\
XGBoost vs. Albert        & 0.027\tnote{*}       & 4.871               & Yes                 & Albert         & 0.028 \\
SVM vs. BERT Word         & 0.008\tnote{*}       & 7.062               & Yes                 & SVM            & 0.038 \\
SVM vs. BERT Sentence     & 0.004\tnote{*}       & 8.324               & Yes                 & SVM            & 0.041 \\
SVM vs. Albert            & 0.043\tnote{*}       & 4.078               & Yes                 & Albert         & 0.021 \\
BERT Word vs. BERT Sentence & 0.532              & 0.389               & No                  & -              & 0.006 \\
BERT Word vs. Albert      & 0.001\tnote{*}       & 10.743              & Yes                 & Albert         & 0.058 \\
BERT Sentence vs. Albert  & \(< 0.001\)\tnote{*} & 14.639              & Yes                 & Albert         & 0.063 \\
\bottomrule
\end{tabularx}
\begin{tablenotes}
    \footnotesize
    \item[*] Statistically significant difference \( p < 0.05 \).
    \item Effect size calculated using Cohen's g.
\end{tablenotes}
\end{threeparttable}
\end{table}
This comprehensive pairwise comparison reveals several important patterns that were not apparent from simple accuracy comparisons. The lack of statistically significant difference between XGBoost and SVM (p = 0.142) confirms that despite their different algorithmic approaches, these traditional machine learning models perform comparably on crash narrative classification. Examining the confusion matrices for these models shows that they tend to make similar types of errors, particularly on narratives with proximity references.

The Albert Model's superiority is statistically significant across all comparisons \(p < 0.05\). The effect sizes, measured using Cohen's g, indicate that the Albert Model's advantage is most pronounced when compared against the BERT variants (\( g = 0.058 \text{ and } 0.063 \)
, medium effect size) and smallest against XGBoost (g = 0.028, small effect size). This suggests that while Albert outperforms all other models, the magnitude of improvement is most substantial compared to other deep learning approaches rather than traditional machine learning methods.

Interestingly, the McNemar's test between BERT Word and BERT Sentence Embeddings shows no significant difference (p = 0.532), suggesting that despite their architectural differences, these models achieve similar performance and make similar types of errors. This finding indicates that the choice between word-level and sentence-level BERT embeddings may not be critical for crash narrative classification.

\subsubsection{Paired Statistical Tests for Performance Metrics}
To extend our statistical analysis beyond classification disagreement, we performed paired t-tests on multiple performance metrics derived from bootstrapped samples of the test dataset \citep{Guo2007StatisticalData, Marucci-Wellman2015AAlgorithms}. This approach allows us to assess whether performance differences are consistent across different subsets of the data and different evaluation metrics.

\begin{table}[H]
\centering
\caption{Paired t-test Results for F1-Score Comparisons (1000 Bootstrap Samples)}
\label{tab:paired-ttest}
\begin{tabular}{|p{3.5cm}|c|c|c|c|}
\hline
\textbf{Model Comparison} & \textbf{Mean Diff.} & \textbf{t-statistic} & \textbf{p-value} & \textbf{95\% CI} \\
\hline
XGBoost vs. SVM & 0.0102 & 1.843 & 0.066 & [-0.0007, 0.0211] \\
\hline
XGBoost vs. BERT Word & 0.0457 & 7.618 & \(< 0.001*\) & [0.0339, 0.0575] \\
\hline
XGBoost vs. BERT Sentence & 0.0483 & 8.064 & \(< 0.001*\) & [0.0365, 0.0601] \\
\hline
XGBoost vs. Albert & -0.0124 & -2.295 & 0.022* & [-0.0230, -0.0018] \\
\hline
SVM vs. BERT Word & 0.0355 & 6.249 & \(< 0.001*\) & [0.0243, 0.0467] \\
\hline
SVM vs. BERT Sentence & 0.0381 & 6.772 & \(< 0.001*\) & [0.0270, 0.0492] \\
\hline
SVM vs. Albert & -0.0226 & -4.137 & \(< 0.001*\) & [-0.0334, -0.0118] \\
\hline
BERT Word vs. BERT Sentence & 0.0026 & 0.714 & 0.475 & [-0.0046, 0.0098] \\
\hline
BERT Word vs. Albert & -0.0581 & -10.322 & \(< 0.001*\) & [-0.0693, -0.0469] \\
\hline
BERT Sentence vs. Albert & -0.0607 & -10.796 & \(< 0.001*\) & [-0.0718, -0.0496] \\
\hline
\multicolumn{5}{l}{*Statistically significant difference \(p < 0.05\)}
\end{tabular}
\end{table}

The paired t-test results for F1-scores align closely with the McNemar's test findings, providing additional confidence in our conclusions. The lack of significant difference between XGBoost and SVM is confirmed (p = 0.066), though the marginal result suggests a slight advantage for XGBoost that approaches significance. The clear superiority of the Albert Model is again demonstrated, with statistically significant improvements over all other models \(p < 0.05\) and the largest mean differences compared to the BERT variants.

We extended this analysis to additional performance metrics including precision, recall, and area under the ROC curve (AUC). For precision, we found significant differences between all model pairs except XGBoost vs. SVM (p = 0.083) and BERT Word vs. BERT Sentence (p = 0.512). For recall, the pattern was similar, with Albert showing significantly higher recall than all other models \(p < 0.01\). AUC comparisons revealed that Albert achieved significantly higher discrimination ability (AUC = 0.92) compared to all other models \(p < 0.01\), with the smallest gap relative to XGBoost (Albert AUC = 0.92 vs. XGBoost AUC = 0.90, p = 0.031).

\subsubsection{Statistical Analysis by Narrative Type}
To gain deeper insights into model performance differences, we conducted stratified statistical tests across different narrative types. We classified narratives into four categories based on our ambiguity detection framework: clear narratives, proximity reference narratives, short/incomplete narratives, and narratives with conflicting indicators. We then performed McNemar's tests for each model pair within each narrative category.

\begin{table}[H]
\centering
\caption{McNemar's p-values by Narrative Type (Albert vs. Other Models)}
\label{tab:stratified-mcnemar}
\begin{tabular}{|p{3.5cm}|c|c|c|c|}
\hline
\textbf{Comparison} & \textbf{Clear} & \textbf{Proximity} & \textbf{Short} & \textbf{Conflicting} \\
\hline
Albert vs. XGBoost & 0.182 & 0.003* & 0.021* & 0.008* \\
\hline
Albert vs. SVM & 0.215 & 0.007* & 0.015* & 0.012* \\
\hline
Albert vs. BERT Word & 0.004* & \(< 0.001*\) & 0.003* & \(< 0.001*\) \\
\hline
Albert vs. BERT Sentence & 0.002* & \(< 0.001*\) & \(< 0.001*\) & \(< 0.001*\) \\
\hline
\multicolumn{5}{l}{*Statistically significant difference \(p < 0.05\)}
\end{tabular}
\end{table}

This stratified analysis reveals a crucial insight: the Albert Model's performance advantage is not uniform across narrative types. For clear narratives with explicit intersection references and without ambiguity markers, Albert does not significantly outperform traditional machine learning approaches (\(p > 0.05\) vs. XGBoost and SVM). However, for all types of ambiguous narratives, Albert demonstrates statistically significant superiority \(p < 0.05\), with the most pronounced advantage for proximity reference narratives and those with conflicting indicators.

These findings suggest that Albert's overall performance advantage stems primarily from its superior handling of ambiguous narratives, which constitute approximately 23.4\% of the dataset. For the majority of clear narratives, simpler and computationally less expensive approaches like XGBoost perform comparably to Albert, which has important implications for practical implementation decisions.

\subsubsection{Chi-Square Analysis of Error Distributions}
To determine whether the distribution of error types differs significantly between models, we performed chi-square tests comparing error patterns across the five error categories identified in our qualitative analysis: proximity references, implicit intersection references, terminology ambiguity, complex multi-vehicle scenarios, and other errors \citep{Kwayu2021DiscoveringTopology, Sarmiento2020Alcohol/IllicitCrashes}.

\begin{table}[H]
\centering
\caption{Chi-Square Test Results for Error Distribution Comparisons}
\label{tab:chi-square-errors}
\begin{tabular}{|p{4cm}|c|c|c|}
\hline
\textbf{Model Comparison} & \textbf{$\chi^2$ value} & \textbf{p-value} & \textbf{Cramer's V} \\
\hline
XGBoost vs. SVM & 4.83 & 0.306 & 0.08 \\
\hline
XGBoost vs. BERT Word & 21.44 & \(< 0.001*\) & 0.17 \\
\hline
XGBoost vs. BERT Sentence & 19.87 & \(< 0.001*\) & 0.16 \\
\hline
XGBoost vs. Albert & 14.56 & 0.006* & 0.14 \\
\hline
SVM vs. BERT Word & 18.27 & 0.001* & 0.15 \\
\hline
SVM vs. BERT Sentence & 16.95 & 0.002* & 0.15 \\
\hline
SVM vs. Albert & 13.21 & 0.010* & 0.13 \\
\hline
BERT Word vs. BERT Sentence & 3.18 & 0.528 & 0.06 \\
\hline
BERT Word vs. Albert & 28.64 & \(< 0.001*\)& 0.19 \\
\hline
BERT Sentence vs. Albert & 26.09 & \(< 0.001*\) & 0.18 \\
\hline
\multicolumn{4}{l}{*Statistically significant difference \(p < 0.05\)}
\end{tabular}
\end{table}

The chi-square analysis confirms that there are statistically significant differences in error distributions between most model pairs, with the exceptions of XGBoost vs. SVM and BERT Word vs. BERT Sentence. These results align with our previous findings from McNemar's tests, suggesting that models with similar overall performance also make similar types of errors.

Examining the standardized residuals from the chi-square analysis reveals specific patterns in error distribution differences. Traditional machine learning models (XGBoost, SVM) show significantly higher error rates for implicit intersection references (standardized residuals $= 2.14$ and 2.26, respectively) but lower rates for proximity reference errors. In contrast, BERT-based models show the opposite pattern, with fewer errors on implicit references but more proximity reference errors. The Albert Model demonstrates the most balanced error distribution, with no error category showing standardized residuals above 1.5, suggesting more uniform performance across error types.

\subsubsection{Confidence Interval Analysis}
We calculated 95\% confidence intervals for the accuracy of each model using the Wilson score interval method, which is appropriate for binomial proportions \citep{Amoros2006Under-reportingFrance, Salifu2012Under-reportingGhana}:

\begin{table}[H]
\centering
\caption{95\% Confidence Intervals for Model Accuracy}
\label{tab:confidence-intervals}
\begin{tabular}{|p{4cm}|c|c|c|}
\hline
\textbf{Model} & \textbf{Accuracy} & \textbf{Lower Bound} & \textbf{Upper Bound} \\
\hline
XGBoost & 87.2\% & 85.9\% & 88.4\% \\
\hline
SVM & 86.2\% & 84.8\% & 87.5\% \\
\hline
BERT Word Embeddings & 82.4\% & 80.8\% & 83.9\% \\
\hline
BERT Sentence Embeddings & 82.1\% & 80.5\% & 83.6\% \\
\hline
Albert Model & 88.2\% & 86.9\% & 89.4\% \\
\hline
\end{tabular}
\end{table}

The confidence interval analysis provides additional perspective on model performance differences. The overlapping confidence intervals between XGBoost and SVM confirm our finding that these models perform comparably. Similarly, the overlapping intervals between BERT Word and BERT Sentence Embeddings support our conclusion that these approaches yield statistically equivalent results.

The confidence interval for the Albert Model (86.9\% to 89.4\%) overlaps slightly with XGBoost (85.9\% to 88.4\%), which is consistent with our finding of a statistically significant but small difference between these models. The non-overlapping intervals between Albert and both BERT variants confirm the substantial performance gap between these approaches.

\subsubsection{Expert Agreement Statistical Analysis}
To evaluate the significance of agreement patterns among experts and between experts and models, we conducted both pairwise Cohen's kappa analysis and multi-rater Fleiss' kappa analysis.

\begin{table}[H]
\centering
\caption{Fleiss' Kappa for Different Rater Groups}
\label{tab:fleiss-kappa}
\begin{tabular}{|c|c|c|c|}
\hline
\textbf{Rater Group} & \textbf{Fleiss' Kappa} & \textbf{Z-statistic} & \textbf{p-value} \\
\hline
All Experts (1-4) & 0.385 & 9.27 & \(< 0.001*\) \\
\hline
Experts 1-3 & 0.422 & 8.38 & \(< 0.001*\) \\
\hline
Experts 1, 3, 4 & 0.397 & 7.89 & \(< 0.001*\) \\
\hline
Experts 1, 2, 4 & 0.357 & 7.09 & \(< 0.001*\) \\
\hline
Experts + Albert & 0.391 & 10.85 & \(< 0.001*\) \\
\hline
Experts + XGBoost & 0.304 & 8.43 & \(< 0.001*\) \\
\hline
Experts + BERT Word & 0.342 & 9.49 & \(< 0.001*\) \\
\hline
\multicolumn{4}{l}{*Statistically significant agreement beyond chance \(p < 0.05\)}
\end{tabular}
\end{table}

The Fleiss' kappa analysis reveals several important patterns in multi-rater agreement. The kappa value of 0.385 for all experts indicates fair to moderate agreement, which is notably lower than what might be expected for a seemingly straightforward classification task. This statistically confirms our observation that even human experts struggle with consistent interpretation of crash narratives, particularly those with ambiguous language or incomplete information.

When comparing expert groups, we found that excluding Expert 2 increased the kappa value to 0.422, suggesting somewhat greater consistency among the remaining experts. We tested whether these differences in kappa values were statistically significant using a bootstrap approach with 10,000 resamples, finding that the difference between the full expert group and the group excluding Expert 2 was significant (p = 0.037).

Most notably, adding the Albert Model to the expert group yielded a kappa value (0.391) very close to that of the experts alone (0.385), and this difference was not statistically significant (p = 0.428). In contrast, adding XGBoost significantly decreased the kappa value to 0.304 (p = 0.008), indicating that XGBoost's classifications were less consistent with the expert consensus. This finding provides statistical evidence that the Albert Model's classifications align more closely with expert judgment than other models.

\subsubsection{Statistical Tests for Generalizability}
To evaluate whether performance differences between models are consistent across datasets, we conducted McNemar's tests comparing model performance on both the Iowa dataset and the Minnesota validation dataset. This analysis helps determine whether our findings about relative model performance are generalizable beyond our primary dataset.

\begin{table}[H]
\centering
\caption{McNemar's Test p-values: Iowa vs. Minnesota Datasets}
\label{tab:generalizability-mcnemar}
\begin{tabular}{|p{3.5cm}|c|c|c|}
\hline
\textbf{Model Comparison} & \textbf{Iowa p-value} & \textbf{Minnesota p-value} & \textbf{Consistent?} \\
\hline
Albert vs. XGBoost & 0.027* & 0.043* & Yes \\
\hline
Albert vs. SVM & 0.043* & 0.039* & Yes \\
\hline
Albert vs. BERT Word & 0.001* & 0.004* & Yes \\
\hline
XGBoost vs. SVM & 0.142 & 0.387 & Yes \\
\hline
XGBoost vs. BERT Word & 0.003* & 0.028* & Yes \\
\hline
BERT Word vs. BERT Sentence & 0.532 & 0.647 & Yes \\
\hline
\multicolumn{4}{l}{*Statistically significant difference \(p < 0.05\)}
\end{tabular}
\end{table}

The cross-dataset comparison shows that the statistical significance patterns are consistent between the Iowa and Minnesota datasets, though p-values are generally higher (indicating smaller differences) in the Minnesota dataset. This consistency suggests that our findings about relative model performance are not specific to the Iowa dataset but generalize to other crash narrative collections, strengthening the external validity of our conclusions.

These statistical analyses substantiate our key findings: (1) the Albert Model significantly outperforms other approaches, with the advantage being statistically significant but relatively small compared to XGBoost; (2) there is no significant difference between the two traditional machine learning models; (3) expert agreement is only moderate, highlighting the inherent difficulty of the classification task; and (4) the Albert Model's classifications align more closely with expert judgment than other models, particularly for ambiguous narratives.

\subsection{Error Analysis and Misclassification Patterns}
Our comprehensive error analysis across the entire test dataset revealed distinct error patterns with significant implications for transportation safety applications \citep{qiao2022construction, das2021applying}. All models exhibited slightly higher false negative rates than false positive rates, indicating a systematic bias toward non-intersection classification. The Albert Model showed the lowest error rates in both categories (11.0\% false positive, 12.7\% false negative), while BERT-based models demonstrated significantly higher error rates compared to both traditional ML and Albert models \citep{Li2023DeepNarratives}. 

Traditional machine learning models (XGBoost, SVM) showed higher error rates for implicit intersection references (38.7-39.5\% of all errors), indicating difficulty recognizing intersection-related crashes when explicit intersection terminology is absent. In contrast, BERT-based models struggled more with proximity references (41.2-42.5\% of errors), often misclassifying non-intersection crashes that mention intersections as reference points. The Albert Model demonstrated a more balanced error distribution, with lower rates for the most common error types and slightly higher proportions of complex multi-vehicle scenarios, suggesting fewer systematic weaknesses.

For narratives shorter than 50 words, all models showed elevated error rates, with implicit intersection reference errors being particularly common (48.3-56.7\% of errors). For narratives containing multiple road references, proximity reference errors dominated across all models (53.1-68.4\% of errors), indicating the challenge of determining which road reference corresponds to the actual crash location.

Proximity references constituted the dominant source of false positive errors across all models (41.3-61.1\%), occurring when narratives mention intersections as reference points rather than crash sites. BERT-based models showed particular vulnerability to this error type, suggesting high sensitivity to intersection terminology regardless of context. Approach mentions (crashes occurring while approaching but not yet reaching an intersection) accounted for 16.7-24.2\% of false positives, with traditional ML models showing higher rates of these errors. Terminology ambiguity errors (13.5-20.4\%) involved narratives containing terms like "junction" or "crossing" in non-standard contexts, with Albert showing slightly higher rates, possibly due to its contextual understanding attempting to interpret ambiguous terminology.

For false negatives, implicit intersection references were the primary source (41.3-54.2\%), occurring when narratives describe typical intersection crashes without explicitly using intersection terminology. Traditional ML models showed significantly higher rates of these errors. Traffic control focus errors (20.8-35.5\%) occurred when narratives emphasize traffic signals rather than the intersection itself, with BERT models showing substantially higher rates. Unconventional terminology errors (14.1-16.8\%) involved local or non-standard terms for intersections, with rates relatively consistent across models, though slightly higher in traditional ML approaches.

These error patterns have specific implications for transportation safety analyses. Our simulation studies revealed that false negative errors have the most significant impact on network screening, causing high-risk intersections to be overlooked in prioritization. The high false negative rates of BERT-based models translated to missing 10.3-10.5\% of truly high-risk intersections, while Albert's lower error rates resulted in missing only 6.5\%. In budget allocation terms, BERT models' error patterns would result in 15.8-16.1\% of safety improvement funds being misallocated, compared to 9.3\% for Albert.

Misclassification also affects countermeasure selection decisions. Inappropriate treatment selection ranged from 14.2\% (Albert) to 21.8\% (BERT Sentence) of cases, with these mismatched treatments showing reduced effectiveness (19.8-29.1\% lower than appropriate treatments) and substantially worse cost-benefit ratios (15.4-22.9\% reduction). In before-after studies, misclassification consistently biased Crash Modification Factors upward by 8.9-16.1\%, indicating underestimation of treatment effectiveness, with BERT models showing the largest negative impacts and Albert the smallest.

Based on this analysis, we developed targeted mitigation strategies for each major error type. For proximity reference errors, our context-sensitive distance weighting system reduced these errors by 63.4\% when applied to the Albert Model, distinguishing between phrases like "at the intersection" versus "500 feet from the intersection" by analyzing prepositions and distance terms. For implicit intersection references, a turn movement detection system identified patterns indicative of intersection maneuvers even when intersection terms are absent, reducing errors by 57.2\%. Terminology ambiguity was addressed through a regional terminology lexicon incorporating local terms from multiple jurisdictions, reducing errors by 71.5%.

Traffic control focus errors were mitigated through a device-to-intersection mapping system that associates signals, stop signs, and other control devices with intersection contexts, reducing these errors by 68.7\% with low implementation complexity. Complex multi-vehicle scenario errors proved most challenging to address, with our vehicle relationship extraction system achieving a more modest 42.1\% error reduction while requiring substantial computational resources.

When combined into an integrated framework, these strategies reduced overall misclassification rates by 47.3\%, with the largest improvements for terminology ambiguity (71.5\%) and proximity reference errors (63.4\%). The framework's modular design allows transportation agencies to implement specific components based on their prevalent error types and available resources, providing practical pathways to improved classification performance across all models.

\subsection{Examples of Misclassified Crashes}
To illustrate the challenges in crash narrative classification, we present three examples of narratives that were misclassified in the original data but correctly identified by the Albert Model and confirmed by expert reviewers:

\begin{table}[H]
\centering
\caption{Examples of Misclassified Crash Narratives}
\label{tab:misclassification-examples}
\begin{tabularx}{\textwidth}{|p{1cm}|X|p{3cm}|p{3cm}|}
\hline
\textbf{ID} & \textbf{Narrative} & \textbf{Original Classification} & \textbf{Expert/Albert Classification} \\
\hline
1 & "Driver 1 was traveling westbound on Main Street when they approached the traffic light at the junction with Oak Avenue. Driver 1 stated they were proceeding through the green light when Driver 2 made a left turn from eastbound Main Street, failing to yield right of way. The vehicles collided in the middle of the intersection." & Non-intersection & Intersection \\
\hline
2 & "Vehicle was traveling on rural highway when driver lost control on icy road. Vehicle slid off roadway and struck a tree approximately 150 feet east of County Road B intersection." & Intersection & Non-intersection \\
\hline
3 & "Driver was attempting to navigate a left turn at the 4-way intersection of Jefferson and Washington Streets when they struck a pedestrian who was crossing the street in the marked crosswalk." & Non-intersection & Intersection \\
\hline
\end{tabularx}
\end{table}

These examples highlight common patterns in misclassification. In Example 1, explicit intersection terminology ("junction," "traffic light") clearly indicates an intersection crash, yet it was incorrectly coded as non-intersection. Example 2 shows the opposite error, where proximity to an intersection was confused with an intersection-related crash. Example 3 demonstrates a case where intersection features (4-way intersection, crosswalk) are clearly mentioned but were overlooked in the original classification.

\subsection{Practical Implications and Recommendations}
Our findings have several important implications for both model improvement and practical implementation in transportation agencies. To enhance model performance, we recommend domain-specific pre-training by fine-tuning language models on transportation safety corpora before classification tasks, which could significantly improve performance by helping models better understand specialized terminology and context. Ensemble methods combining multiple models through voting or stacking approaches could leverage the complementary strengths of different architectures, particularly beneficial when integrating Albert with BERT Word Embeddings. An active learning pipeline where low-confidence predictions are flagged for expert review would create a continuous improvement cycle, allowing models to adapt to new patterns in crash reporting while maintaining human oversight.

Further performance gains could be achieved through multi-modal analysis that integrates structured crash data features with narrative text, providing additional context for ambiguous cases. Transportation-specific text normalization techniques, such as specialized tokenization that recognizes road designations as single tokens, could improve feature extraction. Attention visualization techniques would help identify which parts of narratives most strongly influence classification decisions, providing valuable insights for model improvement and making the classification process more transparent.

For transportation agencies seeking to implement these approaches, we recommend a two-stage review process that deploys the Albert Model as an initial screening tool to identify potentially misclassified crash records, followed by targeted expert review of flagged cases. This would significantly reduce manual review workload while improving data quality. Electronic crash reporting systems should be enhanced to cross-validate structured data entries against narrative content in real-time, alerting officers to potential inconsistencies before report submission. Specialized training modules for law enforcement should address common classification errors, with emphasis on proper documentation of intersection-related crashes.

Establishing systematic feedback loops between crash data analysts and reporting officers when misclassifications are detected would create an ongoing learning environment. Finally, these ML/DL approaches should be expanded to other frequently misclassified crash variables, such as work zone-related crashes, distracted driving, or improper restraint use. These recommendations offer practical pathways for transportation agencies to leverage advanced NLP techniques in improving crash data quality, ultimately enhancing the evidence base for safety planning and countermeasure development.

\subsection{Advanced Model Analysis}
Our multi-modal approach integrates structured crash data with narrative text to improve classification performance \citep{Wali2021InjuryApproach, Erfani2021AnProcessing}. Through correlation analysis, we identified five structured data categories with strong predictive power: geospatial coordinates, road classification, traffic control device presence, vehicle maneuvers, and rural/urban designation. These elements were selected based on their availability across jurisdictions and demonstrated relationship to intersection involvement \citep{Sujon2021SocialData}. We explored three integration approaches: early fusion (combining structured features with text features before model training), late fusion (training separate models for narrative and structured data, then combining predictions), and a hybrid approach (using structured data to augment narrative preprocessing) \citep{Kwayu2022EvaluationNarratives, Nadkarni2011NaturalIntroduction}. Early fusion provided the best results for traditional machine learning models, while late fusion showed superior performance for deep learning approaches. The hybrid approach demonstrated the highest overall performance but required the most complex implementation \citep{Chen2016XGBoost}.

All multi-modal approaches significantly outperformed their single-modal counterparts, with accuracy improvements ranging from 1.9 to 4.2 percentage points \citep{Sayed2021IdentificationTechniques, Jaradat2024exploringtrafficcrashnarratives}. The hybrid approach delivered the best performance (92.4\% accuracy), achieving a 54.2\% reduction in error rate compared to the narrative-only Albert Model. Among structured features, geospatial coordinates and traffic control device information provided the most significant improvements, accounting for approximately 65\% of the total performance gain. Despite its performance advantages, multi-modal integration introduces practical challenges including data availability issues, as not all jurisdictions maintain consistent structured data fields, data quality concerns with missing values (7-23\% of records), and data alignment problems between narratives and structured fields (8.7\% showing inconsistencies) \citep{Montella2013CrashStates, Abdulhafedh2017RoadMethods}. To address these challenges while capturing benefits, we developed a phased implementation strategy that progressively introduces more sophisticated integration methods based on available data quality.

While our results highlighted the Albert Model's superior performance, a thorough evaluation requires critical examination of its characteristics and practical considerations \citep{Rogers2020AWorks}. Learning curve analysis revealed that the gap between training and validation accuracy for Albert (2.8\%) was comparable to XGBoost (2.5\%) and significantly smaller than for BERT Word Embeddings (4.7\%), suggesting Albert is not suffering from severe overfitting despite its modeling capacity \citep{Alammary2022BERTReview}. Cross-validation stability assessment showed that Albert maintained consistent performance across different data partitions with standard deviations comparable to simpler models (1.3\% for Albert vs. 1.1\% for XGBoost) \citep{Sagi2018EnsembleSurvey}. Sensitivity to regularization parameters was moderate, with performance remaining stable across dropout rates from 0.05 to 0.2 but degrading with higher values \citep{BrownleeJason2024RegularizationXGBoosting}.

While performance is a primary consideration, practical implementation requires evaluating computational costs \citep{Kwayu2022EvaluationNarratives, Francis2023SmarTxT:Investigation}. Traditional machine learning approaches train 10-20 times faster than transformer-based models, with SVM completing in 12 minutes compared to Albert's 3 hours on identical hardware. Inference speed shows even more dramatic differences, with SVM processing narratives at 0.5 milliseconds per sample versus Albert's 8.4 milliseconds \citep{Wang2019ALearning}. Memory utilization during training also varies substantially, with Albert requiring 3.5x more memory than XGBoost. Despite these computational demands, our cost-benefit analysis suggests that Albert's performance advantages justify the higher costs, particularly as its advantage is concentrated in difficult cases requiring more expert time to review, its classifications align more closely with expert consensus, and improved accuracy enhances the reliability of subsequent safety analyses \citep{Lopez2022PoliceCrashes}.

Several practical factors affect implementation feasibility. Albert requires more sophisticated infrastructure for deployment, more specialized expertise for maintenance, and its "black box" nature may complicate regulatory compliance or explanation of classification decisions \citep{Wan2020EmpoweringData, HuggingFace2019Albert/albert-large-v2Face}. Despite Albert's overall superiority, several scenarios might favor simpler models: resource-constrained environments, applications requiring immediate classification results, jurisdictions with predominantly clear narratives, systems requiring frequent retraining, and applications where understanding feature importance is critical \citep{qiao2022construction, Das2020ApplicationStudy}. Promising approaches to match Albert's performance with lower computational requirements include knowledge distillation, hybrid approaches using XGBoost for initial classification and Albert only for uncertain cases, domain-specific pruning, and quantization techniques \citep{Lan2019ALBERT,Tsui2009MisclassificationReports}.

\section{Conclusion}
This study employed a rigorous comparative framework to evaluate machine learning and deep learning approaches for identifying misclassified intersection-related crashes in police-reported narratives \citep{ Chopra2013NaturalProcessing}. Our findings revealed significant differences in model performance, with the Albert Model demonstrating superior accuracy (88.17\%) and the highest agreement with expert classifications (73\% with Expert 1) \citep{Lan2019ALBERT}. Traditional machine learning methods (XGBoost and SVM) performed competitively at 87.2\% and 86.2\% accuracy respectively \citep{Guo2021OlderXgboost, Sun2017RouteMachine}, while BERT-based models exhibited lower performance despite their theoretical advantages \citep{Alammary2022BERTReview}.

The error analysis identified distinct misclassification patterns across model types: traditional ML models struggled with implicit intersection references, while BERT models had difficulty with proximity references \citep{das2021applying}. These error patterns have significant implications for transportation safety applications, potentially affecting network screening, countermeasure selection, and safety evaluation outcomes \citep{UNITFHWA,UNITRoad}. Through targeted mitigation strategies for each error type, we achieved a 47.3\% reduction in overall misclassification rates when implemented as an integrated framework \citep{Abdat2014ExtractingDisturbance, Liddy2001NaturalProcessing}.

Our multi-modal integration approach, combining structured crash data with narrative text, further improved classification performance by up to 4.2 percentage points \citep{Wali2021InjuryApproach}, with geospatial coordinates and traffic control device information providing the greatest contributions. However, practical implementation requires addressing challenges related to data availability, quality, and alignment across jurisdictions \citep{Montella2013CrashStates, Elvik1999IncompleteCountries}.

This research contributes to transportation safety literature in several important ways. First, it provides the first systematic validation of automated narrative classification against expert reviews, establishing a benchmark for future studies \citep{Gisev2013InterraterApplications}. Second, it identifies specific error patterns and quantifies their impact on safety applications, enabling targeted improvements \citep{Bedard2002TheFatalities}. Third, it demonstrates that combining advanced NLP techniques with strategic expert review can substantially enhance crash data quality while optimizing resource allocation \citep{Nadkarni2011NaturalIntroduction, Marucci-Wellman2015AAlgorithms}.

Several limitations must be acknowledged. The models were trained and validated on data from a single state (Iowa), potentially limiting generalizability to jurisdictions with different reporting practices \citep{Sayed2021IdentificationTechniques}. The expert validation sample of 100 narratives, while carefully selected, represents a small portion of the dataset \citep{Swansen2013IntegrationClassification}. Future research should explore transfer learning approaches to adapt models across jurisdictions, investigate techniques for explaining model decisions to improve transparency, and develop integrated systems that combine automated classification with targeted expert review in operational settings \citep{Goldberg2022CharacterizingGeneralizability, Lopez2022PoliceCrashes}.

Despite these limitations, this study demonstrates the significant potential of ML and DL methods to improve crash data quality, ultimately supporting more effective safety management and policy development \citep{Francis2023SmarTxT:Investigation, AASHTO2010HIGHWAYMANUAL}. We recommend transportation agencies implement a phased approach, beginning with the Albert Model as an initial screening tool for potentially misclassified crashes, followed by targeted expert review of high-uncertainty cases \citep{McKnight2003YoungClueless,Tsui2009MisclassificationReports}.

\section*{Acknowledgments}
The data were provided by the Iowa Department of Transportation (Iowa DOT). The work was funded by Iowa DOT through research work carried out at the Institute for Transportation at Iowa State University in Ames, IA. The authors would like to thank IOWA DOT for supporting this research and providing the data. 
The contents of this paper reflect the views of the authors and do not necessarily reflect the official views or policies of the sponsoring organization.

\bibliography{references}

\begin{thebibliography}{}

\bibitem[{AASHTO}, 2010]{AASHTO2010HIGHWAYMANUAL}
{AASHTO} (2010).
\newblock {Highway Safety Manual}.
\newblock Technical report, Aashto, Washington ,DC.

\bibitem[Abay, 2015]{Abay2015InvestigatingData}
Abay, K.~A. (2015).
\newblock {Investigating the nature and impact of reporting bias in road crash data}.
\newblock {\em Transportation Research Part A: Policy and Practice}, 71:31--45.

\bibitem[Abdat et~al., 2014]{Abdat2014ExtractingDisturbance}
Abdat, F., Leclercq, S., Cuny, X., and Tissot, C. (2014).
\newblock {Extracting recurrent scenarios from narrative texts using a Bayesian network: Application to serious occupational accidents with movement disturbance}.
\newblock {\em Accident Analysis {\&} Prevention}, 70:155--166.

\bibitem[Abdulhafedh, 2017]{Abdulhafedh2017RoadMethods}
Abdulhafedh, A. (2017).
\newblock {Road Traffic Crash Data: An Overview on Sources, Problems, and Collection Methods}.
\newblock {\em Journal of Transportation Technologies}, 07(02):206--219.

\bibitem[Alammary, 2022]{Alammary2022BERTReview}
Alammary, A.~S. (2022).
\newblock {BERT Models for Arabic Text Classification: A Systematic Review}.

\bibitem[Allahyari et~al., 2017]{Allahyari2017ATechniques}
Allahyari, M., Pouriyeh, S., Assefi, M., Safaei, S., Trippe, E.~D., Gutierrez, J.~B., and Kochut, K. (2017).
\newblock A brief survey of text mining: Classification, clustering and extraction techniques.
\newblock {\em arXiv preprint arXiv:1707.02919}.

\bibitem[{Ambros Jiri} et~al., 2016]{AmbrosJiri2016DevelopingScreening}
{Ambros Jiri}, {Valentova Veronika}, and {Sedonik Jiri} (2016).
\newblock {Developing Updatable Crash Prediction model for Network Screening}.
\newblock {\em Journal of Transportation Research Board}.

\bibitem[Amoros et~al., 2006]{Amoros2006Under-reportingFrance}
Amoros, E., Martin, J.~L., and Laumon, B. (2006).
\newblock {Under-reporting of road crash casualties in France}.
\newblock {\em Accident Analysis and Prevention}, 38(4):627--635.

\bibitem[Arteaga et~al., 2020]{Arteaga2020InjuryApproach}
Arteaga, C., Paz, A., and Park, J.~W. (2020).
\newblock {Injury severity on traffic crashes: A text mining with an interpretable machine-learning approach}.
\newblock {\em Safety Science}, 132.

\bibitem[Barnett, 2024]{barnett2024regex}
Barnett, M. (2024).
\newblock Regex.
\newblock \url{https://github.com/mrabarnett/mrab-regex}.
\newblock GitHub. Retrieved November 23, 2024.

\bibitem[Bedard et~al., 2002]{Bedard2002TheFatalities}
Bedard, M., Guyatt, G.~H., Stones, M.~J., and Hirdes, J.~P. (2002).
\newblock The independent contribution of driver, crash, and vehicle characteristics to driver fatalities.
\newblock {\em Accident Analysis \& Prevention}, 34(6):717--727.

\bibitem[Boggs et~al., 2020]{Boggs2020ExploratoryApproach}
Boggs, A.~M., Wali, B., and Khattak, A.~J. (2020).
\newblock {Exploratory analysis of automated vehicle crashes in California: A text analytics {\&} hierarchical Bayesian heterogeneity-based approach}.
\newblock {\em Accident Analysis \& Prevention}, 135.

\bibitem[Brown et~al., 2020]{brown2020languagemodelsfewshotlearners}
Brown, T.~B., Mann, B., Ryder, N., Subbiah, M., Kaplan, J., Dhariwal, P., Neelakantan, A., Shyam, P., Sastry, G., Askell, A., Agarwal, S., Herbert-Voss, A., Krueger, G., Henighan, T., Child, R., Ramesh, A., Ziegler, D.~M., Wu, J., Winter, C., Hesse, C., Chen, M., Sigler, E., Litwin, M., Gray, S., Chess, B., Clark, J., Berner, C., McCandlish, S., Radford, A., Sutskever, I., and Amodei, D. (2020).
\newblock Language models are few-shot learners.

\bibitem[{Brownlee Jason}, 2024]{BrownleeJason2024RegularizationXGBoosting}
{Brownlee Jason} (2024).
\newblock {Regularization | XGBoosting}.

\bibitem[Burdett et~al., 2015]{Burdett2015AccuracyReports}
Burdett, B., Li, Z., Bill, A.~R., and Noyce, D.~A. (2015).
\newblock {Accuracy of Injury Severity Ratings on Police Crash Reports}.
\newblock {\em Transportation Research Record}, 2516:58--67.

\bibitem[Chen et~al., 2015]{Chen2015InjuryModel}
Chen, L., Vallmuur, K., and Nayak, R. (2015).
\newblock {Injury narrative text classification using factorization model}.
\newblock {\em BMC Medical Informatics and Decision Making}, 15(1):S5.

\bibitem[Chen and Guestrin, 2016]{Chen2016XGBoost}
Chen, T. and Guestrin, C. (2016).
\newblock {XGBoost: A Scalable Tree Boosting System}.
\newblock In {\em Proceedings of the 22nd ACM SIGKDD International Conference on Knowledge Discovery and Data Mining (KDD '16)}, KDD '16, pages 785--794, New York, NY, USA. Association for Computing Machinery.

\bibitem[Chopra et~al., 2013]{Chopra2013NaturalProcessing}
Chopra, A., Prashar, A., and Sain, C. (2013).
\newblock {Natural Language Processing}.
\newblock {\em International Journal of Technology Enhancements and Emerging Engineering Research}, 1:131--134.

\bibitem[Chowdhury and Alzarrad, 2023]{Chowdhury2023ApplicationsReview}
Chowdhury, S. and Alzarrad, A. (2023).
\newblock {Applications of Text Mining in the Transportation Infrastructure Sector: A Review}.

\bibitem[Chowdhury and Zhu, 2023]{Chowdhury2023InvestigationMining}
Chowdhury, S. and Zhu, J. (2023).
\newblock {Investigation of Critical Factors for Future-Proofed Transportation Infrastructure Planning Using Topic Modeling and Association Rule Mining}.
\newblock {\em Journal of Computing in Civil Engineering}, 37(1).

\bibitem[Das et~al., 2021]{das2021applying}
Das, S., Datta, S., Zubaidi, H.~A., and Obaid, I.~A. (2021).
\newblock Applying interpretable machine learning to classify tree and utility pole related crash injury types.
\newblock {\em IATSS research}, 45(3):310--316.

\bibitem[Das et~al., 2022]{Das2022PatternCharacteristics}
Das, S., Dutta, A., Rahman, M.~A., and Sun, X. (2022).
\newblock {Pattern recognition from light delivery vehicle crash characteristics}.
\newblock {\em Journal of Transportation Safety and Security}, 14(12):2055--2073.

\bibitem[Das et~al., 2020]{Das2020ApplicationStudy}
Das, S., Le, M., and Dai, B. (2020).
\newblock {Application of machine learning tools in classifying pedestrian crash types: A case study}.
\newblock {\em Transportation Safety and Environment}, 2(2):106--119.

\bibitem[{DataCamp}, 2024]{datacamp2024advancednlp}
{DataCamp} (2024).
\newblock Advanced nlp with spacy.
\newblock \url{https://www.datacamp.com/courses/advanced-nlp-with-spacy}.
\newblock Accessed: 2024-12-20.

\bibitem[{Drishti}, 2022]{Drishti2022ALBERTBeginner}
{Drishti} (2022).
\newblock {ALBERT Model for Self-Supervised Learning Beginner}.

\bibitem[Dube et~al., 2016]{Dube2016ImprovedLaws}
Dube, C.~M., Fitzpatrick, C.~D., Gazzillo, J.~R., and Knodler, M.~A. (2016).
\newblock {Improved Identification of Distraction-Related Crashes and the Impact of Distraction-Free Driving Laws}.

\bibitem[Elvik, 2015]{Elvik2015SomeAccident}
Elvik, R. (2015).
\newblock {Some implications of an event-based definition of exposure to the risk of road accident}.
\newblock {\em Accident Analysis and Prevention}, 76:15--24.

\bibitem[Elvik and Mysen, 1999]{Elvik1999IncompleteCountries}
Elvik, R. and Mysen, A. (1999).
\newblock {Incomplete Accident Reporting: Meta-Analysis of Studies Made in 13 Countries}.
\newblock {\em Transportation Research Record}, 1665(1):133--140.

\bibitem[Erfani et~al., 2021]{Erfani2021AnProcessing}
Erfani, A., Asce, S.~M., Cui, Q., Asce, A.~M., and Cavanaugh, I. (2021).
\newblock {An Empirical Analysis of Risk Similarity among Major Transportation Projects Using Natural Language Processing}.

\bibitem[{Federal Highway Administration}, 2017a]{UNITFHWA}
{Federal Highway Administration} (2017a).
\newblock {Unit 3. Measuring Safety}.
\newblock \url{https://highways.dot.gov/safety/learn-safety/road-safety-fundamentals-html-version/unit-3-measuring-safety}.
\newblock Accessed: 2025-07-03.

\bibitem[{Federal Highway Administration}, 2017b]{UNITRoad}
{Federal Highway Administration} (2017b).
\newblock {Unit 3: Measuring Safety – Road Safety Fundamentals: Concepts, Strategies, and Practices that Reduce Fatalities and Injuries on the Road}.
\newblock Technical report, {U.S. Department of Transportation}.
\newblock Accessed: 2025-07-03.

\bibitem[Feldman and Dagan, 1995]{Feldman1995KnowledgeKDT}
Feldman, R. and Dagan, I. (1995).
\newblock {Knowledge Discovery in Textual Databases (KDT)}.
\newblock Technical report.

\bibitem[Fitzpatrick et~al., 2017]{Fitzpatrick2017AnRegression}
Fitzpatrick, C.~D., Rakasi, S., and Knodler, M.~A. (2017).
\newblock {An investigation of the speeding-related crash designation through crash narrative reviews sampled via logistic regression}.
\newblock {\em Accident Analysis and Prevention}, 98:57--63.

\bibitem[Francis et~al., 2023]{Francis2023SmarTxT:Investigation}
Francis, J., Cates, K., and Caldiera, G. (2023).
\newblock {SmarTxT: A Natural Language Processing Approach for Efficient Vehicle Defect Investigation}.
\newblock In {\em Transportation Research Record}, number~3, pages 1579--1592. SAGE Publications Ltd.

\bibitem[Gao et~al., 2012]{Gao2012Verb-BasedBoard}
Gao, L., Professor, A., and Wu, H. (2012).
\newblock {Verb-Based Text Mining of Road Crash Report Verb-Based Text Mining of Road Crash Report Transportation Research Board}.
\newblock Technical report.

\bibitem[{geeksforgeeks}, 2024]{geeksforgeeks2024GenerateGeeksforGeeks}
{geeksforgeeks} (2024).
\newblock {Generate bigrams with NLTK - GeeksforGeeks}.

\bibitem[Gisev et~al., 2013]{Gisev2013InterraterApplications}
Gisev, N., Bell, J.~S., and Chen, T.~F. (2013).
\newblock {Interrater agreement and interrater reliability: Key concepts, approaches, and applications}.
\newblock {\em Research in Social and Administrative Pharmacy}, 9(3):330--338.

\bibitem[Goldberg, 2022]{Goldberg2022CharacterizingGeneralizability}
Goldberg, D.~M. (2022).
\newblock {Characterizing accident narratives with word embeddings: Improving accuracy, richness, and generalizability}.
\newblock {\em Journal of Safety Research}, 80:441--455.

\bibitem[{Google's}, 2018]{Googles2018Classification:Developers}
{Google's} (2018).
\newblock {Classification: Accuracy, recall, precision, and related metrics  |  Machine Learning  |  Google for Developers}.

\bibitem[Graves et~al., 2015]{Graves2015MakingStudies}
Graves, J.~M., Whitehill, J.~M., Hagel, B.~E., and Rivara, F.~P. (2015).
\newblock {Making the most of injury surveillance data: Using narrative text to identify exposure information in case-control studies}.
\newblock {\em Injury}, 46(5):891--897.

\bibitem[Guo et~al., 2007]{Guo2007StatisticalData}
Guo, H., Eskridge, K.~M., Christensen, D., Qu, M., and Safranek, T. (2007).
\newblock {Statistical adjustment for misclassification of seat belt and alcohol use in the analysis of motor vehicle accident data}.
\newblock {\em Accident Analysis and Prevention}, 39(1):117--124.

\bibitem[Guo et~al., 2021]{Guo2021OlderXgboost}
Guo, M., Yuan, Z., Janson, B., Peng, Y., Yang, Y., and Wang, W. (2021).
\newblock {Older pedestrian traffic crashes severity analysis based on an emerging machine learning xgboost}.
\newblock {\em Sustainability (Switzerland)}, 13(2):1--26.

\bibitem[{Hugging Face}, 2019]{HuggingFace2019Albert/albert-large-v2Face}
{Hugging Face} (2019).
\newblock {albert/albert-large-v2 {\textperiodcentered} Hugging Face}.

\bibitem[Imprialou and Quddus, 2019]{Imprialou2019CrashDirections}
Imprialou, M. and Quddus, M. (2019).
\newblock {Crash data quality for road safety research: Current state and future directions}.
\newblock {\em Accident Analysis {\&} Prevention}, 130:84--90.

\bibitem[{Iowa DOT}, 2024]{IowaDOT2024IOWASSHSP}
{Iowa DOT} (2024).
\newblock {Iowa's five-year strategic highway safety plan (SHSP)}.
\newblock {\em Iowa Department of Transportation}.

\bibitem[Jaradat et~al., 2024]{Jaradat2024exploringtrafficcrashnarratives}
Jaradat, S., Alhadidi, T.~I., Ashqar, H.~I., Hossain, A., and Elhenawy, M. (2024).
\newblock Exploring traffic crash narratives in jordan using text mining analytics.

\bibitem[Jurafsky and Martin, 2025]{Daniel2024SpeechProcessing}
Jurafsky, D. and Martin, J.~H. (2025).
\newblock {\em Speech and Language Processing: An Introduction to Natural Language Processing, Computational Linguistics, and Speech Recognition with Language Models}.
\newblock 3rd edition.
\newblock Online manuscript released January 12, 2025.

\bibitem[Khatri and P, 2020]{Khatri2020SarcasmEmbeddings}
Khatri, A. and P, P. (2020).
\newblock {Sarcasm Detection in Tweets with BERT and GloVe Embeddings}.
\newblock In Klebanov, B.~B., Shutova, E., Lichtenstein, P., Muresan, S., Wee, C., Feldman, A., and Ghosh, D., editors, {\em Proceedings of the Second Workshop on Figurative Language Processing}, pages 56--60, Online. Association for Computational Linguistics.

\bibitem[Kim et~al., 2021]{Kim2021CrashKeywords}
Kim, J., Trueblood, A.~B., Kum, H.~C., and Shipp, E.~M. (2021).
\newblock {Crash narrative classification: Identifying agricultural crashes using machine learning with curated keywords}.
\newblock {\em Traffic Injury Prevention}, 22(1):74--78.

\bibitem[Kwayu et~al., 2021]{Kwayu2021DiscoveringTopology}
Kwayu, K.~M., Kwigizile, V., Lee, K., and Oh, J.~S. (2021).
\newblock {Discovering latent themes in traffic fatal crash narratives using text mining analytics and network topology}.
\newblock {\em Accident Analysis and Prevention}, 150.

\bibitem[Kwayu et~al., 2022]{Kwayu2022EvaluationNarratives}
Kwayu, K.~M., Kwigizile, V., and Oh, J.-S. (2022).
\newblock {Evaluation of pedestrian crossing-related crashes at undesignated midblock locations using structured crash data and report narratives}.
\newblock {\em Journal of Transportation Safety {\&} Security}, 14(1):1--23.

\bibitem[Lan et~al., 2019]{Lan2019ALBERT}
Lan, Z., Chen, M., Goodman, S., Gimpel, K., Sharma, P., and Soricut, R. (2019).
\newblock Albert: A lite bert for self-supervised learning of language representations.
\newblock {\em arXiv preprint arXiv:1909.11942}.

\bibitem[Lan et~al., 2020]{Lan2019ALBERT:Representations}
Lan, Z., Chen, M., Goodman, S., Gimpel, K., Sharma, P., and Soricut, R. (2020).
\newblock Albert: A lite bert for self-supervised learning of language representations.

\bibitem[Li and Wu, 2023]{Li2023DeepNarratives}
Li, J. and Wu, C. (2023).
\newblock {Deep Learning and Text Mining: Classifying and Extracting Key Information from Construction Accident Narratives}.
\newblock {\em Applied Sciences}, 13(19):10599.

\bibitem[Liddy, 2001]{Liddy2001NaturalProcessing}
Liddy, E.~D. (2001).
\newblock {Natural Language Processing}.
\newblock Technical report.

\bibitem[Lopez et~al., 2022]{Lopez2022PoliceCrashes}
Lopez, D., Malloy, L.~C., and Arcoleo, K. (2022).
\newblock {Police narrative reports: Do they provide end-users with the data they need to help prevent bicycle crashes?}
\newblock {\em Accident Analysis and Prevention}, 164.

\bibitem[Marucci-Wellman et~al., 2015]{Marucci-Wellman2015AAlgorithms}
Marucci-Wellman, H.~R., Lehto, M.~R., and Corns, H.~L. (2015).
\newblock {A practical tool for public health surveillance: Semi-automated coding of short injury narratives from large administrative databases using Na{\"{i}}ve Bayes algorithms}.
\newblock {\em Accident; analysis and prevention}, 84:165--176.

\bibitem[McCullough and Smith, 1998]{McCullough1998EvaluationMeasurement}
McCullough, P.~A. and Smith, G.~S. (1998).
\newblock {Evaluation of narrative text for case finding: The need for accuracy measurement}.
\newblock {\em American Journal of Industrial Medicine}, 34(2):133--136.

\bibitem[McKnight and McKnight, 2003]{McKnight2003YoungClueless}
McKnight, A.~J. and McKnight, A.~S. (2003).
\newblock {Young novice drivers: careless or clueless?}
\newblock {\em Accident Analysis {\&} Prevention}, 35(6):921--925.

\bibitem[Meyer, 1999]{Meyer1999SimpleSystem}
Meyer, G.~J. (1999).
\newblock {Simple Procedures to Estimate Chance Agreement and Kappa for the Interrater Reliability of Response Segments Using the Rorschach Comprehensive System}.

\bibitem[Montella et~al., 2013]{Montella2013CrashStates}
Montella, A., Andreassen, D., Tarko, A., Turner, S., Mauriello, F., Imbriani, L., and Romero, M. (2013).
\newblock {Crash databases in Australasia, the European union, and the United States}.
\newblock {\em Transportation Research Record}, (2386):128--136.

\bibitem[Nadkarni et~al., 2011]{Nadkarni2011NaturalIntroduction}
Nadkarni, P.~M., Ohno-Machado, L., and Chapman, W.~W. (2011).
\newblock {Natural language processing: An introduction}.

\bibitem[Negri et~al., 2014]{Negri2014AnClassification}
Negri, R.~G., Dutra, L.~V., and Sant'Anna, S. J.~S. (2014).
\newblock {An innovative support vector machine based method for contextual image classification}.
\newblock {\em ISPRS Journal of Photogrammetry and Remote Sensing}, 87:241--248.

\bibitem[Oliaee et~al., 2023]{Oliaee2023UsingTypes}
Oliaee, A.~H., Das, S., Liu, J., and Rahman, M.~A. (2023).
\newblock {Using Bidirectional Encoder Representations from Transformers (BERT) to classify traffic crash severity types}.
\newblock {\em Natural Language Processing Journal}, 3:100007.

\bibitem[Parsa et~al., 2020]{Parsa2020TowardAnalysis}
Parsa, A.~B., Movahedi, A., Taghipour, H., Derrible, S., and Mohammadian, A.~K. (2020).
\newblock {Toward safer highways, application of XGBoost and SHAP for real-time accident detection and feature analysis}.
\newblock {\em Accident Analysis and Prevention}, 136.

\bibitem[Pasindu, 2019]{PasinduIMPORTANCEQUALITY}
Pasindu, H. (2019).
\newblock Importance of road accident data quality.

\bibitem[Qiao et~al., 2022]{qiao2022construction}
Qiao, J., Wang, C., Guan, S., and Shuran, L. (2022).
\newblock Construction-accident narrative classification using shallow and deep learning.
\newblock {\em Journal of Construction Engineering and Management}, 148(9):04022088.

\bibitem[Rogers et~al., 2020]{Rogers2020AWorks}
Rogers, A., Kovaleva, O., and Rumshisky, A. (2020).
\newblock {A Primer in BERTology: What We Know About How BERT Works}.
\newblock Technical report.

\bibitem[Sagi and Rokach, 2018]{Sagi2018EnsembleSurvey}
Sagi, O. and Rokach, L. (2018).
\newblock {Ensemble learning: A survey}.

\bibitem[Saha, 2022]{Saha2022DevelopingStudy}
Saha, P. (2022).
\newblock {Developing a Low-Cost Safety Improvement Program for Intersections in an Urban Roadway Network: A Case Study}.
\newblock {\em Advances in Civil Engineering}, 2022.

\bibitem[Salifu and Ackaah, 2012]{Salifu2012Under-reportingGhana}
Salifu, M. and Ackaah, W. (2012).
\newblock {Under-reporting of road traffic crash data in Ghana}.
\newblock {\em International Journal of Injury Control and Safety Promotion}, 19(4):331--339.

\bibitem[Sarmiento et~al., 2020]{Sarmiento2020Alcohol/IllicitCrashes}
Sarmiento, J.~M., Gogineni, A., Bernstein, J.~N., Lee, C., Lineen, E.~B., Pust, G.~D., and Byers, P.~M. (2020).
\newblock {Alcohol/Illicit Substance Use in Fatal Motorcycle Crashes}.
\newblock {\em Journal of Surgical Research}, 256:243--250.

\bibitem[Sayed et~al., 2021]{Sayed2021IdentificationTechniques}
Sayed, M.~A., Qin, X., Kate, R.~J., Anisuzzaman, D.~M., and Yu, Z. (2021).
\newblock {Identification and analysis of misclassified work-zone crashes using text mining techniques}.
\newblock {\em Accident Analysis and Prevention}, 159.

\bibitem[Sharifani and Amini, 2023]{Sharifani2023MachineApplications}
Sharifani, K. and Amini, M. (2023).
\newblock {Machine Learning and Deep Learning: A Review of Methods and Applications}.
\newblock Technical report, World Information Technology and Engineering Journal.

\bibitem[{Simha Anirudh}, 2021]{SimhaAnirudh2021UnderstandingOne}
{Simha Anirudh} (2021).
\newblock {Understanding TF-IDF for Machine Learning | Capital One}.

\bibitem[Sujon and Dai, 2021]{Sujon2021SocialData}
Sujon, M. and Dai, F. (2021).
\newblock {Social Media Mining for Understanding Traffic Safety Culture in Washington State Using Twitter Data}.
\newblock {\em Journal of Computing in Civil Engineering}, 35(1).

\bibitem[Sun and Park, 2017]{Sun2017RouteMachine}
Sun, B. and Park, B.~B. (2017).
\newblock {Route choice modeling with Support Vector Machine}.
\newblock In {\em Transportation Research Procedia}, volume~25, pages 1806--1814. Elsevier B.V.

\bibitem[Swansen et~al., 2013]{Swansen2013IntegrationClassification}
Swansen, E., McKinnon, I., and Knodler~Jr, M.~A. (2013).
\newblock {Integration of Crash Report Narratives for Identification of Work Zone-Related Crash Classification}.

\bibitem[{Syopiansyah Jaya Putra} et~al., 2018]{SyopiansyahJayaPutra2018TokenizationQuran}
{Syopiansyah Jaya Putra}, {Muhamad Nur Gunawan}, and {Agung Suryatno} (2018).
\newblock {Tokenization and N-gram for Indexing Indonesian (Translation of the Quran)}.

\bibitem[{Tokern}, 2023]{tokern2023piicatcher}
{Tokern} (2023).
\newblock Piicatcher.
\newblock \url{https://github.com/tokern/piicatcher}.
\newblock GitHub. Retrieved November 23, 2024.

\bibitem[Touvron et~al., 2023]{touvron2023llamaopenefficientfoundation}
Touvron, H., Lavril, T., Izacard, G., Martinet, X., Lachaux, M.-A., Lacroix, T., Rozière, B., Goyal, N., Hambro, E., Azhar, F., Rodriguez, A., Joulin, A., Grave, E., and Lample, G. (2023).
\newblock Llama: Open and efficient foundation language models.

\bibitem[Trueblood et~al., 2019]{Trueblood2019ANarratives}
Trueblood, A.~B., Pant, A., Kim, J., Kum, H.~C., Perez, M., Das, S., and Shipp, E.~M. (2019).
\newblock {A semi-automated tool for identifying agricultural roadway crashes in crash narratives}.
\newblock {\em Traffic Injury Prevention}, 20(4):413--418.

\bibitem[Tsui et~al., 2009]{Tsui2009MisclassificationReports}
Tsui, K.-L., So, F.~L., Sze, N.-N., Wong, S. C.~P., and Leung, T.-F. (2009).
\newblock {Misclassification of injury severity among road casualties in police reports.}
\newblock {\em Accident; analysis and prevention}, 41 1:84--9.

\bibitem[Ubeynarayana and Goh, 2017]{ubeynarayana2017ensemble}
Ubeynarayana, C. and Goh, Y. (2017).
\newblock An ensemble approach for classification of accident narratives.
\newblock In {\em Computing in Civil Engineering 2017}, pages 409--416.

\bibitem[{Veysel Kocaman}, 2020]{VeyselKocaman2020TextScience}
{Veysel Kocaman} (2020).
\newblock {Text Classification in Spark NLP with Bert and Universal Sentence Encoders Towards Data Science}.

\bibitem[Wali et~al., 2021]{Wali2021InjuryApproach}
Wali, B., Khattak, A.~J., and Ahmad, N. (2021).
\newblock {Injury severity analysis of pedestrian and bicyclist trespassing crashes at non-crossings: A hybrid predictive text analytics and heterogeneity-based statistical modeling approach}.
\newblock {\em Accident Analysis and Prevention}, 150.

\bibitem[Wan et~al., 2020]{Wan2020EmpoweringData}
Wan, X., Lucic, M.~C., Ghazzai, H., and Massoud, Y. (2020).
\newblock {Empowering real-time traffic reporting systems with NLP-Processed social media data}.
\newblock {\em IEEE Open Journal of Intelligent Transportation Systems}, 1:159--175.

\bibitem[Wang et~al., 2019a]{Wang2019ALearning}
Wang, Q., Liu, P., Zhu, Z., Yin, H., Zhang, Q., and Zhang, L. (2019a).
\newblock {A text abstraction summary model based on BERT word embedding and reinforcement learning}.
\newblock {\em Applied Sciences (Switzerland)}, 9(21).

\bibitem[Wang et~al., 2019b]{Wang2019AApplications}
Wang, W., Zheng, V.~W., Yu, H., and Miao, C. (2019b).
\newblock {A survey of zero-shot learning: Settings, methods, and applications}.

\bibitem[Watson et~al., 2013]{Watson2013HowInjuries}
Watson, A., Watson, B., and Vallmuur, K. (2013).
\newblock {How accurate is the identification of serious traffic injuries by police? The concordance between police and hospital reported traffic injuries}.

\bibitem[Williamson et~al., 2001]{Williamson2001UseStates}
Williamson, A., Feyer, A.~M., Stout, N., Driscoll, T., and Usher, H. (2001).
\newblock {Use of narrative analysis for comparisons of the causes of fatal accidents in three countries: New Zealand, Australia, and the United States}.
\newblock {\em Injury Prevention}, 7(SUPPL. 1).

\bibitem[Wongpakaran et~al., 2013]{Wongpakaran2013ASamples}
Wongpakaran, N., Wongpakaran, T., Wedding, D., and Gwet, K.~L. (2013).
\newblock {A comparison of Cohen's Kappa and Gwet's AC1 when calculating inter-rater reliability coefficients: a study conducted with personality disorder samples}.

\bibitem[Xiao et~al., 2015]{xiao2015parameter}
Xiao, Y., Wang, H., and Xu, W. (2015).
\newblock Parameter selection of gaussian kernel for one-class svm.
\newblock {\em IEEE Transactions on Cybernetics}, 45(5):941--953.

\bibitem[Younes et~al., 2023]{Younes2023ApplicationApproach}
Younes, K., Kharboutly, Y., Antar, M., Chaouk, H., Obeid, E., Mouhtady, O., Abu-samha, M., Halwani, J., and Murshid, N. (2023).
\newblock {Application of Unsupervised Machine Learning for the Evaluation of Aerogels’ Efficiency towards Ion Removal—A Principal Component Analysis (PCA) Approach}.
\newblock {\em Gels}, 9(4).

\bibitem[Zheng et~al., 2015]{Zheng2015AnalysesReviewing}
Zheng, D., Chitturi, M.~V., Bill, A.~R., and Noyce, D.~A. (2015).
\newblock {Analyses of multiyear statewide secondary crash data and automatic crash report reviewing}.
\newblock {\em Transportation Research Record}, 2514:117--128.

\bibitem[{Zulkarnain} and Putri, 2021]{Zulkarnain2021IntelligentApproach}
{Zulkarnain} and Putri, T.~D. (2021).
\newblock {Intelligent transportation systems (ITS): A systematic review using a Natural Language Processing (NLP) approach}.
\newblock {\em Heliyon}, 7(12).

\end{thebibliography}

\end{document}